# An Efficient Consolidation of Word Embedding and Deep Learning Techniques for Classifying Anticancer Peptides: FastText+BiLSTM

Onur Karakaya[1] and Zeynep Hilal Kilimci[2]

[1]Research and Development Inc, Turkcell Technology, Istanbul, 34854, Türkiye.
[2]Department of Information Systems Engineering, Kocaeli University, Kocaeli, 41001, Türkiye.

Contributing author: zeynep.kilimci@kocaeli.edu.tr

**Abstract**

Anticancer peptides (ACPs) are a group of peptides that exhibit antineoplastic properties. The utilization of ACPs in cancer prevention can present a viable substitute for conventional cancer therapeutics, as they possess a higher degree of selectivity and safety. Recent scientific advancements generate an interest in peptide-based therapies which offer the advantage of efficiently treating intended cells without negatively impacting normal cells. However, as the number of peptide sequences continues to increase rapidly, developing a reliable and precise prediction model becomes a challenging task. In this work, our motivation is to advance an efficient model for categorizing anticancer peptides employing the consolidation of word embedding and deep learning models. First, Word2Vec and FastText are evaluated as word embedding techniques for the purpose of extracting peptide sequences. Then, the output of word embedding models are fed into deep learning approaches CNN, LSTM, BiLSTM. To demonstrate the contribution of proposed framework, extensive experiments are carried on widely-used datasets in the literature, ACPs250 and Independent. Experiment results show the usage of proposed model enhances classification accuracy when compared to the state-of-the-art studies. The proposed combination, FastText+BiLSTM, exhibits 92.50% of accuracy for ACPs250 dataset, and 96.15% of accuracy for Independent dataset, thence determining new state-of-the-art.

## 1 Introduction

Cancer, a devastating disease characterized by uncontrolled cell growth and metastasis, continues to be a leading cause of death worldwide. According to the World Health Organization, various types of cancer cause millions of deaths annually, and this number is increasing each year. According to recent reports, the global incidence of newly diagnosed cancer cases surpassed 19.3 million cases, resulting in approximately 10 million deaths in 2020, as per the available data [1]. The COVID-19 pandemic has resulted in significant disruptions to cancer diagnosis and treatment due to the closure of healthcare facilities, employment and health insurance disruptions, and concerns over COVID-19 exposure. While the most substantial impact was observed during the peak of the pandemic in mid-2020, the healthcare system has not fully recovered. Notably, the Massachusetts General Hospital experienced a reduction in surgical oncology procedures, with levels reaching only 72% of those in 2019 during the latter half of 2020 and showing a modest recovery to 84% in 2021, representing the lowest rebound among all surgical specialties [2]. The introduction of delays in both the diagnosis and treatment of cancer can contribute to an increase in the prevalence of advanced-stage disease and subsequent mortality rates [3].

Traditional cancer treatments such as chemotherapy and radiation therapy often have limitations in terms of effectiveness, selectivity, and adverse side effects. Although conventional anticancer therapies, including chemotherapy and radiation, exhibit high efficacy, they are accompanied by considerable costs and pose undesired side effects on healthy cells. Furthermore, cancer cells can develop resistance mechanisms against conventional chemotherapeutic agents [4]. Conventional treatments like radiation therapy, targeted therapy, and chemotherapy often yield suboptimal results and can cause numerous side effects in recipients. Some of these side effects include cognitive impairment, insomnia, gastrointestinal problems, compromised immune system, thrombocytopenia, and anemia [5, 6]. Therefore, there is an urgent need for more effective treatments to overcome the mentioned disadvantages.

In recent years, peptide-based therapy has emerged as a promising approach for cancer treatment. Anticancer peptides (ACPs) have shown promise as candidates due to their ability to selectively target and disrupt cancer cell membranes, interfere with intracellular processes, and induce apoptosis. Anticancer peptides (ACPs) typically refer to short peptide fragments derived from protein sequences, characterized by containing fewer than 50 amino acids [7]. Presently, diverse peptide-based therapeutic approaches have been employed in clinical and preclinical trials for the treatment of various types of tumors [8][9]. The identification and classification of ACPs traditionally rely on experimental techniques such as high-throughput screening and mass spectrometry. However, these approaches are time-consuming, labor-intensive, and limited by the availability of peptide libraries. To overcome these challenges, computational methods have emerged as valuable tools for predicting and classifying ACPs [10].

In the late 1980s, neural networks garnered significant attention within the field of Machine Learning (ML) and Artificial Intelligence (AI) as a result of the emergence of several proficient learning algorithms and network architectures [11]. During that period, groundbreaking methods such as multilayer perceptron networks trained using "Backpropagation" algorithms, self-organizing maps, and radial basis function networks emerged as innovative approaches [12–14]. Although neural networks have demonstrated successful applications in numerous domains, research interest in the field waned over time. However, in 2006, a pivotal advancement called "Deep Learning" (DL) was introduced by Hinton et al. (2006) [15], which was founded on the principles of artificial neural networks (ANNs). The advent of deep learning sparked a resurgence in neural network research, often referred to as "new-generation neural networks." This resurgence can be attributed to the remarkable achievements of deep networks in effectively addressing a wide range of classification and regression tasks [11]. Nowadays, deep learning (DL) technology has emerged as a prominent subject in the domains of machine learning, artificial intelligence (AI), data science, and analytics. This can be attributed to its remarkable ability to learn from provided data. Within the realm of its operational domain, DL is regarded as a subset of both ML and AI, representing an AI function that emulates the cognitive data processing capabilities of the human brain. Notably, DL distinguishes itself from traditional ML by demonstrating enhanced efficiency with increasing data volumes. Leveraging multiple layers to capture data abstractions, DL constructs computational models that facilitate intricate learning processes. Although DL requires extensive training time due to a substantial number of parameters, its testing phase exhibits noteworthy efficiency compared to other ML algorithms, necessitating less computational time [16].In recent years, Machine learning (ML) has found application in various domains, encompassing text mining, spam detection, video recommendation, image classification, and multimedia concept retrieval [17–22]. Deep learning (DL), among the different ML algorithms, has emerged as a widely adopted technique in these applications [23–25]. DL is also known as representation learning (RL). The burgeoning interest in the fields of deep and distributed learning can be attributed to the exponential growth in data availability and the remarkable advancements in hardware technologies, such as High-Performance Computing (HPC) [26]. DL builds upon the foundations of conventional neural networks but significantly surpasses its predecessors in performance. Moreover, DL integrates transformations and graph-based methodologies to construct complex, multi-layer learning models. Recent advancements in DL techniques have demonstrated exceptional performance across diverse applications, including audio and speech processing, visual data processing, and natural language processing (NLP) [27–30].Deep learning, a subset of machine learning, has revolutionized the field of bioinformatics by leveraging multi-layered artificial neural networks to extract complex patterns and representations from complex biological data. The hierarchical architecture of deep neural networks enables them to learn meaningful features directly from raw inputs without the need for manual feature engineering. As a result, deep learning models have shown tremendous potential for analyzing protein sequences and predicting their functional properties.

In this study, an efficient model for classifying anticancer peptides utilizing the consolidation of word embedding techniques and deep learning models is introduced. The motivation of the proposed approach is to evaluate and compare the performance of state-of-the-art deep learning models that aim to achieve high accuracy in distinguishing between anticancer peptides and non-anticancer peptides. To demonstrate the contribution of proposed framework, two different publicly available benchmark datasets namely, ACPs250, and Independent are employed to ensure a fair comparison with the state-of-the-art studies. Furthermore, the impact of different vector models, namely word2Vec, FastText, is investigated to get robust classification results. Experiments results demonstrate that the inclusion of word embedding models when combined with deep learning architectures significantly improves the classification accuracy for detecting anticancer peptides.

The rest of this paper is designated as follows: Section 2 gives a summary of related work on the classification of anticancer peptides. Section 3 introduces the models used in this work. Section 4 describes the proposed framework. Section 5 and Section 6 present experiment results, and conclusions, respectively.

## 2 Related Work

In this section, a brief literature studies on anti-cancer peptide classification is presented.

In [31], a novel multi-channel convolutional neural network (CNN) is proposed for the identification of anticancer peptides (ACPs) from protein sequences. The data from state-of-the-art methodologies is collected and subjected to binary encoding for data preprocessing. Additionally, k-fold cross-validation is utilized to train the models on benchmark datasets, and the performance of the models is compared on independent datasets.

In [32], a novel and highly efficient method, called ME-ACP, is introduced, employing multi-view neural networks combined with an ensemble model for the identification of anticancer peptides. Initially, residue-level and peptide-level features are incorporated using ensemble models based on lightGBMs to yield preliminary results. Subsequently, the outputs of these lightGBM classifiers are fed into a hybrid deep neural network (HDNN) to accurately identify ACPs.

In [33], the construction of expansive non-repetitive training and independent datasets for the study of anticancer peptides is undertaken. Through the utilization of the training dataset, an extensive exploration of diverse feature encodings is conducted, leading to the development of corresponding models employing seven distinct conventional classifiers. Subsequently, a subset of encoding-based models is carefully chosen for each classifier based on their performance metrics. The predicted scores obtained from these selected models are concatenated and further trained using a convolutional neural network (CNN). The resulting predictor, known as MLACP 2.0, is thus established.

In [34], a cutting-edge deep learning model named AntiMF is introduced, which incorporates a multi-view mechanism based on diverse feature extraction models. Comparative analysis reveals that the proposed model outperforms existing methods in predicting anticancer peptides. By employing an ensemble learning framework to extract representations, AntiMF effectively captures multidimensional information, thereby enhancing the comprehensiveness of model representation.

In [35], a novel method called ACP-2DCNN is introduced, which leverages deep learning techniques to enhance the prediction accuracy of anticancer peptides. During the training and prediction stages, the extraction of crucial features is achieved through Dipeptide Deviation from Expected Mean (DDE), while the Two-dimensional Convolutional Neural Network (2D CNN) is employed.

In [36], a deep learning method named MPMABP, based on convolutional neural network (CNN) and bi-directional long short-term memory (BiLSTM), is proposed for the recognition of multi-activities exhibited by bioactive peptides. The MPMABP model employs a stacked architecture consisting of five CNNs operating at different scales, while leveraging the residual network to retain crucial information throughout the learning process. Empirical findings demonstrate the superiority of MPMABP over existing state-of-the-art methods. Analysis of the amino acid distribution reveals a preference for lysine in anti-cancer peptides, leucine in anti-diabetic peptides, and proline in anti-hypertensive peptides.

In [37], a robust framework is developed to accurately identify ACPs. The approach employed incorporates four distinct hypothetical feature encoding mechanisms, namely amino acid, dipeptide, tripeptide, and an enhanced version of pseudo amino acid composition, which effectively capture the motif characteristics of the target class. Additionally, principal component analysis (PCA) is employed for feature selection, focusing on identifying optimal, deep, and highly variable features. Given the diverse nature of the learning process, a range of algorithms is employed to determine the optimal operational method. Through empirical investigations, it is observed that the support vector machine with hybrid feature space demonstrates superior performance. The proposed framework achieves an accuracy of 97.09% and 98.25% on the benchmark and independent datasets, respectively.

In [38], the utilization of three distinct nature encoding schemes is employed to acquire features from peptide sequences. Among these schemes, particular emphasis is placed on the K-space amino acid pair (KSAAP) method, which effectively captures highly correlated and informative descriptors. In addition to sequential features, composite physiochemical properties are employed to capture localized structural descriptors. Furthermore, the autocovariance technique is utilized to represent the intrinsic residue information of amino acids. To ensure the selection of highly discriminative features and reduce the dimensionality of the proposed descriptors, a novel two-level feature selection (2LFS) method is applied. Finally, various learning hypotheses are explored to identify the most superior operational engine and evaluate the performance of the proposed model. To assess the generalization capability of the model, two diverse benchmark datasets are employed.

In [39], a novel multi-headed deep convolutional neural network model, named ACP-MHCNN, is proposed for the extraction and integration of discriminative features from various information sources in an interactive manner. The model adeptly captures sequence, physicochemical, and evolutionary-based features for the identification of anticancer peptides (ACPs) using diverse numerical representations of peptides while effectively managing parameter overhead. Through rigorous experiments involving cross-validation and an independent dataset, it is demonstrated that ACP-MHCNN significantly outperforms other models in terms of ACP identification on the employed benchmarks. ACP-MHCNN achieves a remarkable improvement over the state-of-the-art model, surpassing it by 6.3% in accuracy, 8.6% in sensitivity, 3.7% in specificity, 4.0% in precision, and 0.20 in Matthews correlation coefficient (MCC).to predict anticancer peptides

In [40], a deep learning approach based on convolutional neural networks is proposed for the prediction of biological activity (EC50, LC50, IC50, and LD50) against six tumor cell types: breast, colon, cervix, lung, skin, and prostate. It is demonstrated that models generated through multitask learning exhibit superior performance compared to traditional single-task models. In repeated 5-fold cross-validation using the CancerPPD dataset, the best models, defined within the applicability domain, achieve an average mean squared error of 0.1758, a Pearson's correlation coefficient of 0.8086, and a Kendall's correlation coefficient of 0.6156. To enhance interpretability, the contribution of each residue in the peptide sequence to the predicted activity is inferred by analyzing feature importance weights obtained from the convolutional layers of the model. This novel method, known as xDeep-AcPEP, offers valuable insights for the identification of effective ACPs in the rational design of therapeutic peptides.

In [41], a sequence-based model for the identification of ACPs is proposed, referred to as SAP (Sequence-based ACP Predictor). In SAP, the peptide is characterized by either 400D features or 400D features with g-gap dipeptide features. Subsequently, the unrelated features are selectively eliminated through the utilization of the maximum relevance-maximum distance method.

In [42], an effective computational model is proposed for the prediction of therapeutic peptides (PTPD) utilizing deep learning and word2vec techniques. The representation vectors of all k-mers are obtained through word2vec, incorporating the co-existence information among k-mers. Subsequently, the original peptide sequences are partitioned into k-mers using the windowing method. The embedding vector acquired from word2vec is employed to map the peptide sequences to the input layer. The construction of feature maps involves the application of three types of filters in the convolutional layers, along with dropout and max-pooling operations. These feature maps are then concatenated into a fully connected dense layer, incorporating rectified linear units (ReLU) and dropout operations to prevent overfitting of PTPD. The classification probabilities are generated by employing a sigmoid function. The performance of PTPD is assessed on two distinct datasets: an independent anticancer peptide dataset and a virulent protein dataset, yielding accuracies of 96% and 94% respectively.

In [43], a novel computational model named ACPGCN is introduced, which leverages graph convolution networks to enable the automatic and precise prediction of anticancer peptides (ACPs). In this model, the prediction of ACPs is framed as a graph classification task, wherein each peptide sample is transformed into a graph representation. This approach represents a pioneering step in considering the graph-based nature of ACP prediction.

In [44], a comprehensive examination is undertaken on three fundamental deep learning architectures, namely convolutional, recurrent, and convolutional-recurrent networks, to discern between anticancer peptides (ACPs) and non-ACPs. It is observed that the recurrent neural network featuring bidirectional long short-term memory cells surpasses other architectural designs in performance. Through the utilization of the proposed model, a sequence-based deep learning tool (DeepACP) is developed to effectively assess the probability of a peptide manifesting anticancer activity.

The study [45] introduces ACPNet, a novel deep learning-based model specifically designed for discriminating between anticancer peptides and non-anticancer peptides (non-ACPs). ACPNet incorporates three distinct sources of peptide sequence information, including peptide physicochemical properties and auto-encoding features, which are integrated into the model's training process. ACPNet utilizes a hybrid architecture that combines fully connected networks with recurrent neural networks, capitalizing on the unique strengths of each approach. Evaluation of ACPNet on the ACPs82 dataset reveals notable improvements, such as a 1.2% increase in Accuracy, a 2.0% enhancement in F1-score, a 7.2% boost in Recall, and well-balanced results in terms of the Matthews correlation coefficient.

The study [46] leverages recent advancements in machine learning to present a groundbreaking deep neural network classifier tailored for the prediction of anticancer peptide activity using sequence information. The classifier exhibits performance, as evidenced by its cross-validated accuracy of 98.3%, a Matthews correlation coefficient of 0.91, and an Area Under the Curve (AUC) value of 0.95.

In [47], a word embedding approach based on FastText has been utilized to represent each peptide sample through the use of a skip-gram model. By extracting the descriptors of peptide embeddings, the deep neural network (DNN) model has been applied to effectively discriminate the antimicrobial cyclic peptides (ACPs). The DNN model's parameters have been optimized, resulting in impressive accuracies of 96.94%, 93.41%, and 94.02% when utilizing training, alternate, and independent samples, respectively. Notably, the proposed cACP-DeepGram model has demonstrated superior performance, exhibiting approximately 10% higher prediction accuracy compared to existing predictors. This suggests that the cACP-DeepGram model holds great promise as a reliable tool for scientists and can significantly contribute to academic research and drug discovery endeavors.

## 3  Models

This section offers an extensive exposition of word embedding models and deep learning methodologies that have been proposed and advanced specifically for the task of ACP classification.

### 3.1  Word2Vec Word Embedding Model (Word2Vec)

Word2Vec [48] is a word embedding model widely employed in natural language processing. It operates by generating dense vector representations for words in a continuous vector space. The model takes input text, typically a large corpus of documents, and learns to assign vector embeddings to words based on their contextual usage.

Word2Vec comprises multiple layers, including an input layer, hidden layer(s), and an output layer. The input layer accepts word sequences from the corpus, while the hidden layer(s) learn to capture the contextual relationships between words. The output layer predicts the probability of neighboring words given a target word. The model incorporates various parameters that can be adjusted, such as the dimensionality of the word vectors, the window size (i.e., the number of adjacent words considered for context), and the number of negative samples employed during training.

Word2Vec's working logic revolves around optimizing the word vectors by iteratively predicting neighboring words given a target word using techniques like skip-gram or continuous bag-of-words (CBOW). The model updates the word vectors using stochastic gradient descent optimization during training. In Fig. 1, the architecture of Word2Vec training models are presented.

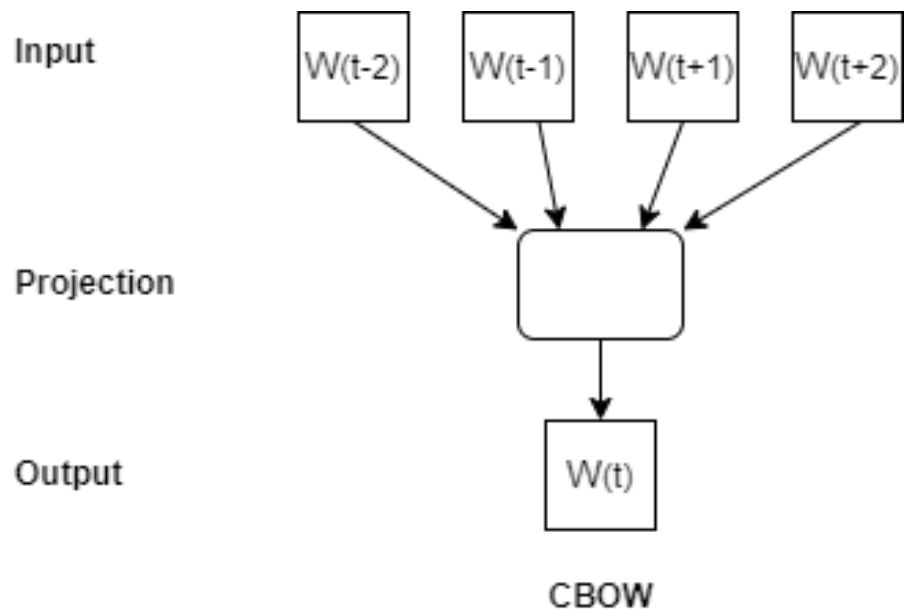
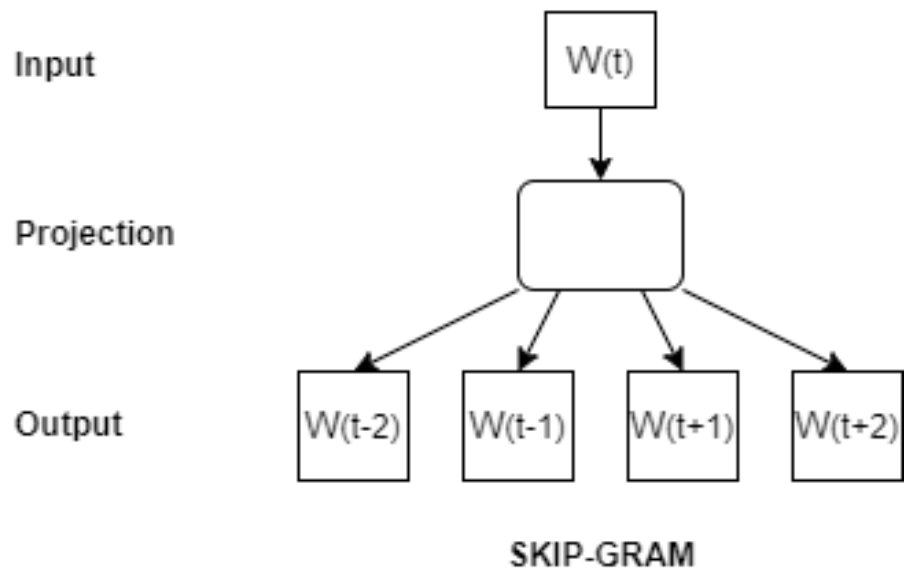

**Fig. 1** The architecture of Word2Vec models

## 3.2 FastText Word Embedding Model (FastText)

FastText [49] is a word embedding model developed for natural language processing tasks. It operates by representing words as continuous-valued vectors, capturing semantic and syntactic information. FastText takes input text data and processes it to learn word embeddings in a unsupervised manner. It uses a shallow neural network architecture with multiple layers to achieve efficient and effective word representations.

The model consists of an input layer, followed by a hidden layer and an output layer. In the input layer, text data is transformed into n-gram character sequences, which capture subword information. These subword sequences are then passed through the hidden layer, which applies a non-linear transformation to generate feature representations. Finally, the output layer predicts the context of the word based on the learned representations. By incorporating subword information, FastText is able to capture morphological variations and handle out-of-vocabulary words more effectively compared to traditional word embeddings.

FastText model has various parameters that affect its performance. The most important one is the embedding dimension, which determines the size of the word vectors. Additionally, the model allows adjusting the learning rate, the number of epochs for training, and the size of the context window. In Fig. 2, the architecture of FastText training model are presented.

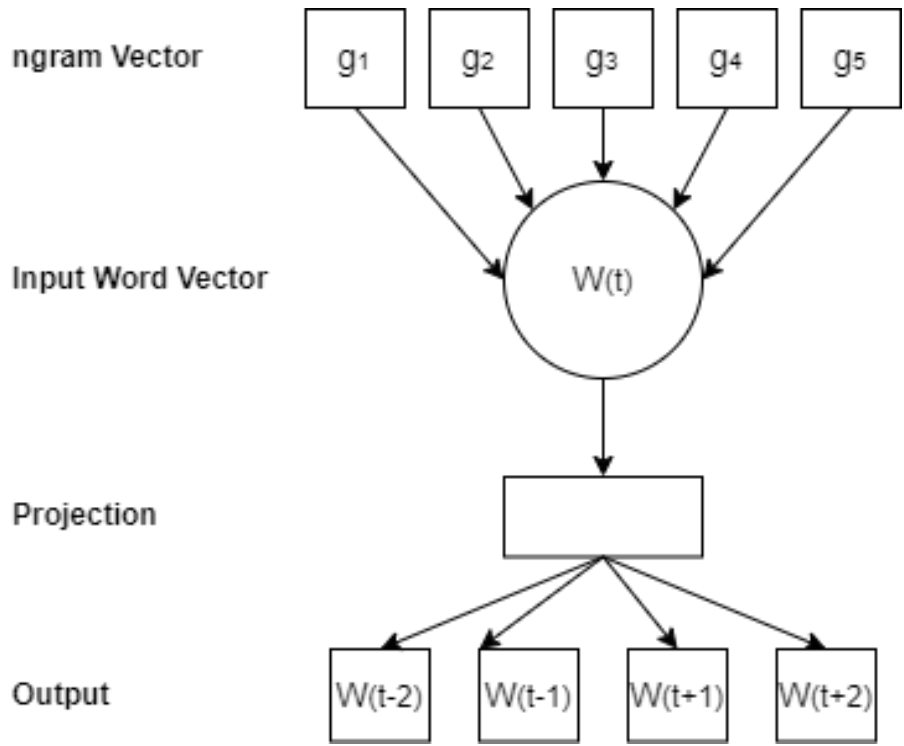

**Fig. 2** The architecture of FastText model

## 3.3 Convolutional Neural Network (CNN)

CNN [50] is a deep learning model that operates by leveraging convolutional layers to extract features from input data, particularly in the domain of image and text analysis. CNN has demonstrated exceptional capabilities in various domains such as image classification, object detection, and image segmentation, sentiment analysis, document classification, and text generation. It follows a hierarchical architecture that includes multiple layers, each performing specific operations to learn and recognize patterns.

The CNN takes input text or image data and processes it through convolutional layers. These layers apply convolution operations on localized regions of the input, known as receptive fields or filters, to extract relevant features. The filters slide over the input, computing dot products between their weights and the input values, producing feature maps that capture different aspects of the data. The CNN typically consists of an input layer, convolutional layers, pooling layers, fully connected layers, and an output layer. In the convolutional layers, feature extraction is performed using filters with shared weights, enabling the network to learn spatial hierarchies of features. Pooling layers reduce the spatial dimensions of the feature maps, retaining the most salient information. The fully connected layers integrate the extracted features and make predictions, while the output layer provides the final classification or regression output.

CNNs have various parameters that affect their performance, including the number and size of filters, the stride of the convolutional operation, the size of the pooling regions, and the activation functions used. Additionally, hyperparameters such as learning rate, regularization techniques, and optimization algorithms play a crucial role in training the CNN, effectively. The operational principles of convolutional neural networks (CNNs) are grounded in the notion of local receptive fields and shared weights. By applying filters across the input data, CNNs are capable of automatically learning and identifying relevant features. The subsequent pooling layers help in reducing the dimensionality of the extracted features while retaining the most discriminative information. The fully connected layers further combine these features and provide the final output predictions. The fundamental architecture of CNN is referred in Fig. 3.

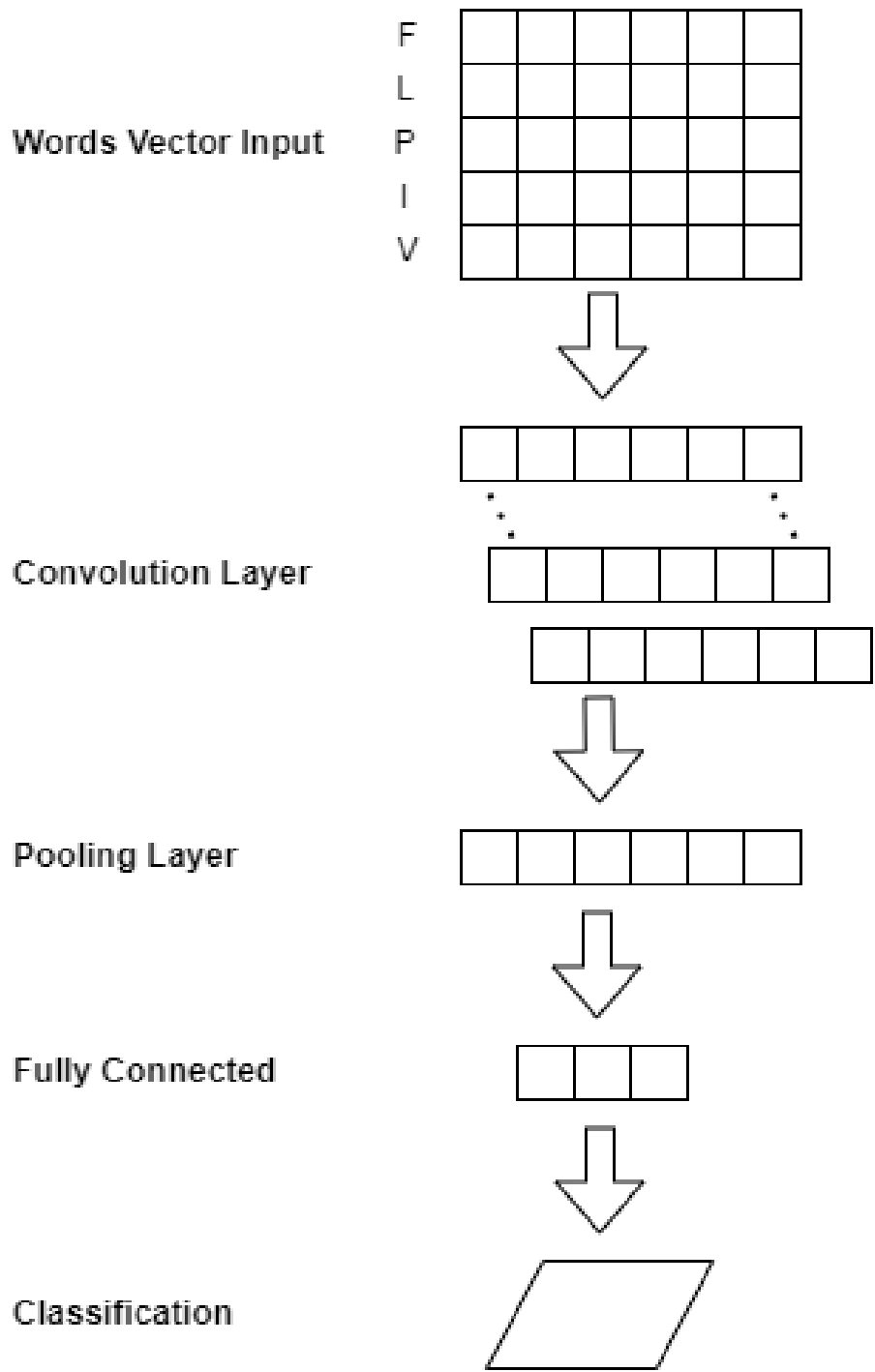

**Fig. 3** The basic CNN architecture

## 3.4 Long short term memory network (LSTM)

LSTM deep learning model is a type of recurrent neural network (RNN) that is designed to overcome the limitations of traditional RNNs in capturing and retaining long-term dependencies in sequential data. It is introduced by Hochreiter and Schmidhuber [51].

LSTMs work by incorporating a memory cell that allows them to selectively retain or forget information over extended sequences. This memory cell is responsible for storing and propagating information across time steps, enabling the network to capture long-term dependencies. The key idea behind LSTMs is the use of gating mechanisms, such as input, forget, and output gates, which regulate the flow of information within the network. LSTMs take sequential data as input, which can be in the form of text, time series, or any other sequential data. Each element of the sequence is typically represented as a vector, and the LSTM processes the sequence one element at a time. LSTMs consist of multiple layers, with each layer containing multiple LSTM units. Each LSTM unit has three main components: the input gate, the forget gate, and the output gate. These gates control the flow of information and the memory cell updates within the LSTM unit. The parameters of an LSTM model include weights and biases associated with each gate and the memory cell. These parameters are learned during the training process using backpropagation through time (BPTT) or other optimization algorithms, where the objective is to minimize a defined loss function.

The performance of LSTM models depends on various factors, such as the size and architecture of the network, the quality and size of the training data, the choice of optimization algorithm, and the hyperparameters of the model. LSTMs have been successfully applied in various domains, including natural language processing, speech recognition, machine translation, and time series analysis. The foundational architecture of LSTM is alluded in Fig. 4.

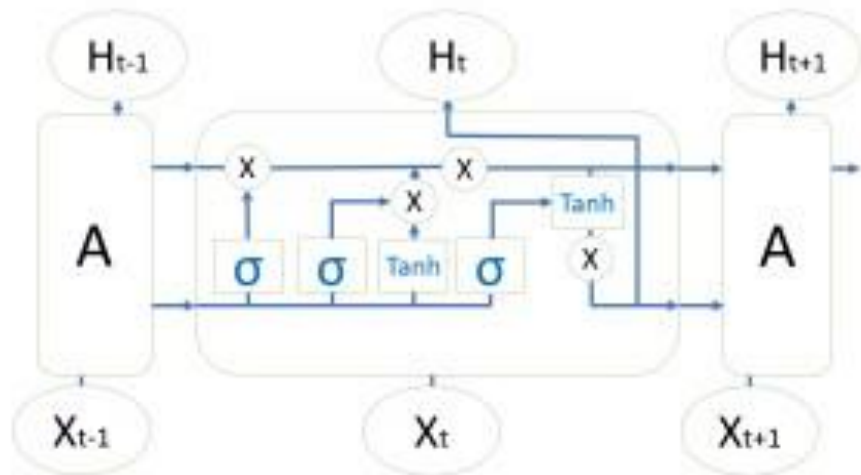

**Fig. 4** The basic LSTM architecture [52]

## 3.5 Bidirectional Long short term memory networks (BiLSTM)

BiLSTM is a deep learning model that operates on sequential data and incorporates both past and future information [53]. It is an extension of the traditional LSTM model and has gained significant popularity for its ability to capture context from both directions. BiLSTM takes input text in the form of sequential data, such as sentences or time series. It is commonly used for tasks such as sequence labeling, sentiment analysis, machine translation, and speech recognition. The performance of the BiLSTM model can be further improved by incorporating techniques like attention mechanisms or additional layers. The input text is typically tokenized into individual units, such as words or characters, and then encoded as numerical vectors using techniques like one-hot encoding or word embeddings.

The BiLSTM model consists of multiple layers of LSTM units. Each LSTM unit has a memory cell and three main gates: the input gate, the forget gate, and the output gate. These gates control the flow of information and regulate the memory cell's state. By incorporating both forward and backward LSTM layers, the BiLSTM model captures dependencies in the input sequence from both directions, allowing it to capture long-term dependencies and context more effectively. The parameters of the BiLSTM model include the number of LSTM units, the number of layers, the input and output dimensions, and the activation functions. The model is trained using backpropagation through time, where the gradients are computed and used to update the weights of the LSTM units. The fundamental architecture of BiLSTM is referred in Fig. 5.

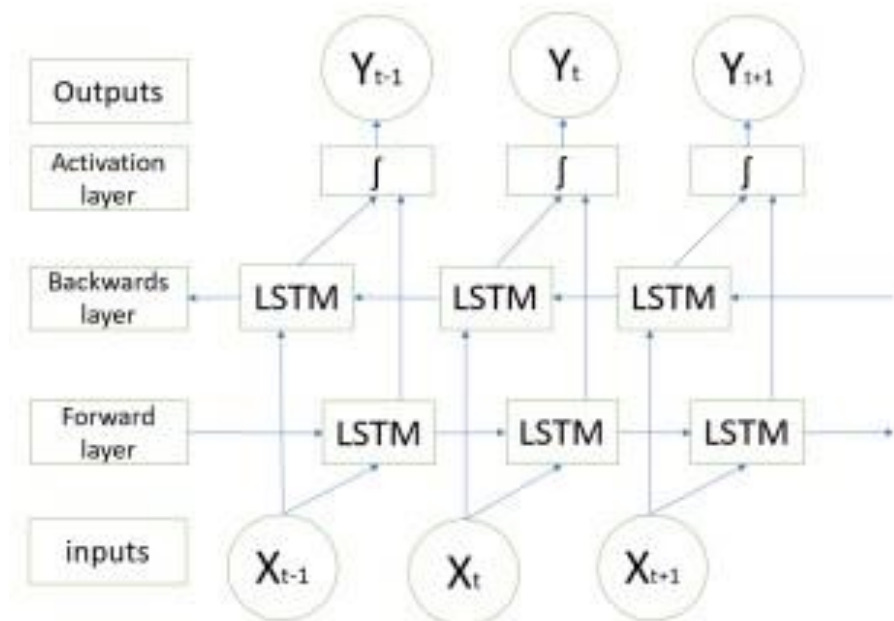

**Fig. 5** The basic BiLSTM architecture [54]

# 4 Proposed Framework

In this section, datasets employed in this study and proposed methodology are introduced.

## 4.1 Dataset

The ACPs250 dataset [55], utilized as one of the primary data sources in this study, comprises a total of 500 peptide instances. Specifically, it encompasses 250 peptides labeled as "anticancer" and an equal number of peptides labeled as "non-anticancer". This balanced dataset plays a crucial role in training and evaluating classification models to discern between these two distinct classes. The inclusion of 250 anticancerlabeled peptides and 250 non-anticancer-labeled peptides allows for a comprehensive analysis of their characteristic features and facilitates the development of effective predictive models. By leveraging the ACPs250 dataset, this study aims to contribute to the field of anticancer peptide research by exploring the discriminative patterns and underlying properties that distinguish anticancer peptides from their non-anticancer counterparts. The statistics and content of the datasets for ACPs250 and Independent datasets are given in Table 1, Table 2, and Table 3.

**Table 1** Statistics of the Datasets

| Dataset | Positive | Negative | Total |
| --- | --- | --- | --- |
| ACPs250 | 250 | 250 | 500 |
| Independent | 150 | 150 | 300 |

**Table 2** Content of the ACPs250 Dataset

| Sample | Content | Label |
| --- | --- | --- |
| 1 | KWKLFKKIEKVGQNIRDGIIKAGPAVA | 0 |
| 2 | FLPAIVGAAAKFLPKIFCAISKKC | 0 |
| ... | ... | ... |
| 499 | FLPIVTNLLSGLL | 1 |
| 500 | GALRGCWTKSYPPKPCK | 1 |

The Independent dataset [56], consisting of 300 peptide sequences, serves as another critical data source in this research endeavor. It encompasses 150 peptides labeled as "anticancer" and an equal number of peptides labeled as "non-anticancer". This dataset offers a unique and separate collection of peptides, distinct from the ACPs250 dataset, to validate and evaluate the generalization capability of the developed models. The inclusion of 150 anticancer-labeled peptides and 150 nonanticancer-labeled peptides in the Independent dataset enables a comprehensive assessment of the model's performance on unseen data, thereby providing insights into its robustness and reliability. By incorporating the Independent dataset, this study aims to strengthen the credibility and applicability of the proposed models in the realm of anticancer peptide prediction and classification. In Table 1, and Table 3, the details of Independent dataset are presented.

Table 3 Content of the Independent Dataset

| Sample | Content | Label |
|---|---|---|
| 1 | AAKKWAKAKWAKAKKWAKAA | 0 |
| 2 | AAVPIVNLKDELLFPSWEALFSGSE | 0 |
| ... | ... | ... |
| 299 | VTSWSLCTPGCTSPGGGSNCSFCC | 1 |
| 300 | YVPLPNVPQPGRRPFPTFPGQG | 1 |

## 4.2 Proposed Methodology

In this study, an efficient classification framework is proposed for the purpose of classifying anticancer peptides with the aid of word embedding models and deep learning methodologies. After gathering widely applied datasets, character-based feature vectors are extracted from peptide sequences on ACPs250, and Independent datasets employing two versions of Word2Vec, and FastText models. The Word2Vec model encompasses both skip-gram and CBOW techniques to perform the word embedding process. For the FastText model, various n-gram units ranging from 2-gram to 4-gram are considered during the embedding process. After constructing feature vector of each sample on both dataset, the processed datasets are fed into CNN, LSTM, and BiLSTM deep learning algorithms. The general flow chart of the proposed framework is given in Fig. 6. As seen in Fig. 6, final decision is ensured by the combination of FastText and BiLSTM model as a consequences of extensive experiments. The architecture of consolidated FastText+BiLSTM framework is presented in Fig. 7.

During the model creation phase, different parameters are utilized. Layer number is varied between 1 and 5, enabling the exploration of models with different layer configurations. Dropout regularization is applied at the rates of 0.1, 0.2, 0.3, 0.4, and 0.5 to prevent over-fitting problem. Batch size values of 16, 32, 64, 96, 128, and 192 are also tested. Some models include max pooling and batch normalization to boost the classification success. Various activation functions, including sigmoid, softmax, and relu, are examined individually or in combination. Different learning rates of 0.1, 0.01, 0.001, and 0.0001 are employed to avoid the issue of over-fitting. Loss types, namely binarycrossentropy and categorical-crossentropy, are employed to achieve optimal results in the classification task. Through the incorporation of these diverse parameters, the performance of the models is thoroughly analyzed. As a result, the most successful models are identified for each dataset.

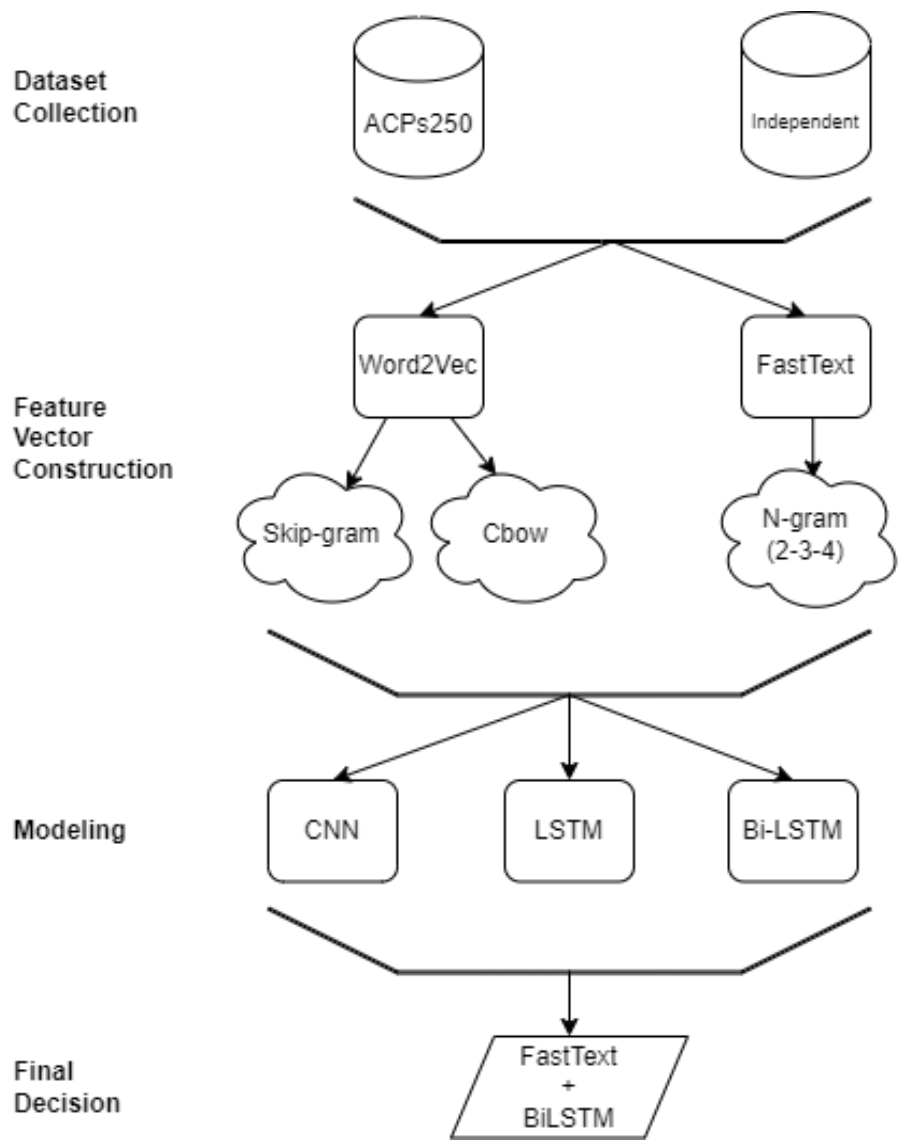

**Fig. 6** A general flow chart of the proposed model.

In details, the CNN model consists of several layers and parameters designed to effectively capture and classify the peptide sequences. The CNN architecture begins with an embedding layer that employs Word2Vec and FastText methods, separately in order to convert peptide sequences into a vector space. This layer maps peptide sequences to a continuous vector space, enabling the capturing of semantic meaning and relationships. Following the embedding layer, the model utilizes a 64-unit CNN layer with a rectified linear unit (ReLU) activation function. This layer performs convolutions on the embedded peptide vectors, extracting relevant features and patterns. Subsequently, dropout regularization with a rate of 0.3 is applied twice to mitigate over-fitting issue. Dropout randomly deactivates a portion of the neurons during training, promoting model generalization. Finally, a softmax activation function is employed in the dense layer to produce probabilistic predictions. This layer computes the probability distribution over the two classes ("anticancer" and "non-anticancer") based on the learned features. The softmax activation ensures that the predicted probabilities sum up to 1, enabling class probabilities interpretation. The model is compiled with the binary-crossentropy loss function, suitable for binary classification tasks, and the Adam optimizer with a learning rate of 0.01. The model is compiled with the binary-crossentropy loss function, which measures the difference between the predicted probabilities and the true labels. Furthermore, the Adam optimizer with a learning rate of 0.01 is used to update the model's weights and biases during the training process. Adam optimizer adapts the learning rate dynamically based on the estimated gradients, facilitating efficient convergence and optimization.

The LSTM model starts with an embedding layer that utilizes Word2Vec and FastText methods, separately for the purpose of transforming the peptide sequences into continuous vector representations, thence, capturing their semantic meaning and contextual information. Following the embedding layer, a LSTM layer with 32 units and a rectified linear unit (ReLU) activation function is utilized. LSTM is a type of recurrent neural network (RNN) that is well-suited for capturing long-term dependencies in sequential data. The LSTM layer processes the embedded peptide vectors, retaining important temporal information. To prevent over-fitting and enhance generalization, two dropout layers with a dropout rate of 0.3 are incorporated into the model. Dropout randomly deactivates a portion of the neurons during training, encouraging the model to rely on more diverse and robust features. Finally, a softmax activation function is employed in the dense layer to generate class probabilities for the classification task. The model is compiled with the binary-crossentropy loss function, suitable for binary classification problems, and the Adam optimizer with a learning rate of 0.01.

Next, a BiLSTM layer with 64 units is employed. This layer is responsible for capturing the sequential dependencies in the text data by processing the input in both forward and backward directions. The bidirectional nature of the BiLSTM allows it to consider the context from both past and future tokens, enabling it to capture longterm dependencies, effectively. Following the BiLSTM layer, a dropout regularization technique is applied with a rate of 0.2. Subsequently, a BiLSTM layer with 32 units is added, followed by another dropout layer with a rate of 0.2. This architecture allows the model to learn more complex relationships and capture finer details in the text data. Finally, a BiLSTM layer with 16 units and a dropout layer with a rate of 0.2 are included. These layers further refine the model's ability to capture intricate patterns and subtle nuances present in the text data. The model concludes with a dense layer utilizing the sigmoid activation function. This layer provides the final output and predicts the class probabilities (e.g., "anticancer" or "non-anticancer"). The model is trained using the binary-crossentropy loss function to minimize the discrepancy between the predicted class probabilities and the true class labels. The Adam optimizer is employed to update the model's weights during training. It adapts the learning rate (set to 0.01) based on the gradient information, allowing the model to converge, efficiently. Overall, this model utilizes a series of BiLSTM layers, dropout regularization, and the embedding layers to effectively capture the complex relationships and temporal dependencies present in text data.

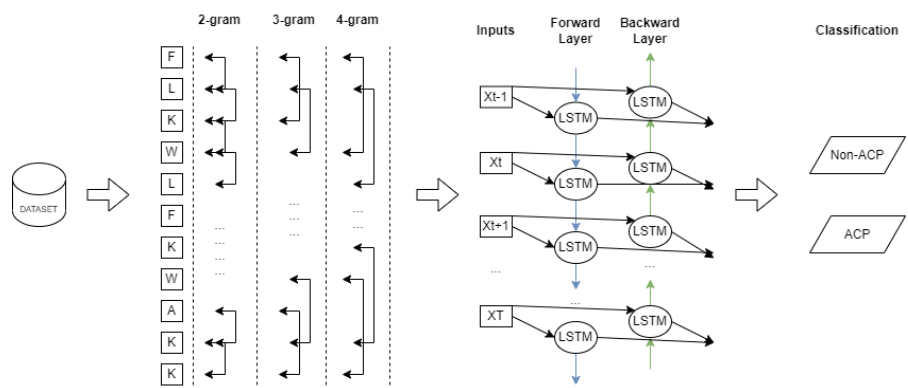

**Fig. 7** The architecture of consolidated FastText+BiLSTM framework.

## 5  Experiment Results

In this section, the consolidation of word embedding models and deep learning techniques are evaluated in terms of different evaluation and performance metrics for the purpose of anticancer peptides classification task. ACC for accuracy, SEN for sensitivity, SPE for specificity, Matthew correlation coefficient for MCC, and area under curve for AUC are abbreviated as evaluation metrics in the tables to demonstrate the classification success of each model and the contribution of our study. These values are given in the table as % ratio. All models are run on Google Colab environment provided free GPU usage by Google. Experiments are accomplished using 90% training and 20% test of data with repeated holdout method. It is applied 10 times on each data set. The following abbreviations are employed for the combination of word embedding techniques and deep learning methods: WS+CNN: Combination of Skip-gram version of Word2Vec and CNN, WC+CNN: Combination of CBOW version of Word2Vec and CNN, WS+LSTM: Combination of Skip-gram version of Word2Vec and LSTM, WC+LSTM: Combination of CBOW version of Word2Vec and LSTM, WS+BiLSTM: Combination of Skip-gram version of Word2Vec and BiLSTM, WC+BiLSTM: Combination of CBOW version of Word2Vec and BiLSTM, FT+CNN: Combination of FastText and CNN, FT+LSTM: Combination of FastText and LSTM, FT+BiLSTM: Combination of FastText and BiLSTM. FastText word embbedding model employed in the experiments are asssesed by varying number of n-gram in the range of 2-4. The best accuracy percentages are represented in bold letters.

As a first attempt, we analyse the effect of all aforementioned models on two different datasets in terms of various evaluation metrics.

**Table 4** Experiment Results of Word2Vec Embedding Model with Deep Learning Algorithms for ACPs250 Dataset

| MODELS | ACC | SEN | SPE | MCC | AUC |
|---|---|---|---|---|---|
| WS+CNN | 57.99 | 59.95 | 57.04 | 16.03 | 62.37 |
| WC+CNN | 74.00 | 72.34 | 79.56 | 41.89 | 81.23 |
| WS+LSTM | 75.99 | 72.68 | 81.23 | 50.96 | 85.79 |
| WC+LSTM | 73.00 | 71.42 | 74.23 | 47.78 | 81.06 |
| WS+BiLSTM | 87.50 | 86.12 | 89.67 | 76.13 | 93.45 |
| WC+BiLSTM | **90.00** | 92.50 | 87.50 | 79.08 | 94.25 |

In Table 4, it is observed that the model employing the WS embedding algorithm, in conjunction with the CNN deep learning architecture, manifests the lowest performance with an accuracy of 57.99%. On the other hand, the model utilizing the WC embedding algorithm and CNN architecture demonstrates a relatively higher accuracy value of 74.00%. Conversely, the integration of the WS embedding algorithm with the LSTM architecture yields a superior accuracy value of 75.99%, while the utilization of the WC with the LSTM architecture results in a slightly lower accuracy value of 73.00%. Notably, the consolidation of WS embedding algorithm and BiLSTM exhibits a superior performance with accuracy of 87.50%. Furthermore, the model utilizing the WC with the BiLSTM architecture showcases the best classification success exhibiting 90.00% of accuracy.

It is clearly observed that the WS embedding algorithm delivers relatively diminished performance in CNN and BiLSTM models when compared to the WC word embedding model. However, in the case of the LSTM model, the WS yields a higher accuracy value when compared to WC version of Word2Vec model. Conversely, the CBOW method attains a higher accuracy value in CNN and BiLSTM models, while exhibiting relatively lower performance in LSTM method. This finding indicates that the WC technique generally possesses a greater ability to capture general word relationships. In conclusion, considering the accuracy values and the employed model types, the fusion of the WC embedding algorithm with the BiLSTM model attains the highest accuracy result.

In Table 5, it is seen that the model combination of WS embedding algorithm with the CNN architecture perform the lowest accuracy score with 57.81%. In comparison, the model employing the WC embedding algorithm with the CNN architecture achieves a significantly higher accuracy value of 84.25%. This suggests that the WC method is more effective in capturing word relationships and performing well when combined with the CNN model. Moving on to the LSTM model, the combination of the WS embedding algorithm with the LSTM architecture yields an accuracy value of 88.75%, which is higher than the WS + CNN model. Similarly, the model utilizing the WC embedding algorithm with the LSTM architecture achieves an accuracy value of 86.51%. These results indicate that both version of Word2Vec embedding algorithms perform well in the LSTM models compared to the CNN method combinations. WS approach exhibiting slightly higher accuracy in comparison with WC model when consolidated with LSTM architecture. Considering the BiLSTM models, the model incorporating the WS embedding algorithm achieves an accuracy value of 89.06%, while the model utilizing the WC embedding algorithm achieves the highest accuracy value of 92.31%. These findings suggest that the combination of the WC embedding algorithm with the BiLSTM architecture yields the best performance among all the models examined.

**Table 5** Experiment Results of Word2Vec Embedding Model with Deep Learning Algorithms for Independent Dataset

| MODELS | ACC | SEN | SPE | MCC | AUC |
|---|---|---|---|---|---|
| WS+CNN | 57.81 | 55.29 | 60.33 | 14.17 | 57.98 |
| WC+CNN | 84.25 | 80.10 | 68.36 | 52.47 | 83.05 |
| WS+LSTM | 88.75 | 89.25 | 88.31 | 77.48 | 93.65 |
| WC+LSTM | 86.51 | 82.62 | 90.74 | 72.83 | 91.25 |
| WS+BiLSTM | 89.06 | 88.88 | 89.37 | 77.65 | 94.05 |
| WC+BiLSTM | **92.31** | 93.79 | 91.52 | 84.62 | 96.55 |

In summary, the analysis of the provided table reveals that the choice of Word2Vec embedding algorithm and deep learning architecture significantly impacts the accuracy of the models. The WC embedding algorithm tends to perform better in the CNN models, while both WS and WC approaches demonstrate good performance in the LSTM models. In terms of the BiLSTM models, the WC embedding algorithm showcases the highest accuracy, indicating its effectiveness in capturing complex relationships in sequential data. As a result of Table 4 and Table 5, the usage of BiLSTM architecture remarkably excels the classification performance when consolidated with CBOW version of Word2Vec model.

**Table 6** Experiment Results of FastText Embedding Model including Various N-grams with Deep Learning Algorithms for ACPs250 Dataset

| MODELS | ACC | SEN | SPE | MCC | AUC |
|---|---|---|---|---|---|
| FT(2)+CNN | 88.99 | 86.27 | 89.75 | 77.63 | 93.51 |
| FT(3)+CNN | 83.99 | 80.58 | 87.42 | 67.29 | 90.20 |
| FT(4)+CNN | 57.99 | 55.79 | 59.83 | 15.74 | 60.29 |
| FT(2)+LSTM | 72.00 | 65.22 | 78.13 | 40.85 | 76.68 |
| FT(3)+LSTM | 75.99 | 70.34 | 82.49 | 49.32 | 82.17 |
| FT(4)+LSTM | 60.22 | 55.79 | 59.83 | 15.74 | 60.29 |
| FT(2)+BiLSTM | 82.99 | 79.68 | 86.34 | 70.01 | 90.01 |
| FT(3)+BiLSTM | **92.50** | 96.29 | 82.60 | 80.27 | 89.45 |
| FT(4)+BiLSTM | 66.15 | 55.79 | 59.83 | 15.74 | 60.29 |

Upon examining the accuracy results of Table 6, it is observed that the model utilizing the FT(2) embedding algorithm with the CNN architecture achieves an accuracy of 88.99%. It is followed by the model incorporating the FT(3) embedding algorithm with the CNN architecture demonstrates a slightly lower accuracy value of 83.99%. The model employing the FT(4) embedding algorithm with the CNN architecture shows the lowest accuracy score with 57.99%. Moving on to the LSTM models, the combination of the FT(2) embedding algorithm with the LSTM architecture attains an accuracy of 72.00%. Additionally, the model utilizing the FT(3) embedding algorithm with the LSTM architecture achieves a higher accuracy value of 75.99%. However, the model incorporating the FT(4) embedding algorithm with the LSTM architecture shows the lowest accuracy value with 60.22%. Lastly, considering the BiLSTM models, the model utilizing the FT(2) embedding algorithm achieves an accuracy value of 82.99%. The combination of the FT(3) embedding algorithm with the BiLSTM architecture yields the highest accuracy with 92.50%. Conversely, the model incorporating the FT(4) embedding algorithm with the BiLSTM architecture demonstrates the lowest accuracy performance with 66.15% similar to the combination of FT(4) and other deep learning architectures.

Analyzing the results, it can be inferred that the choice of FastText n-gram embedding algorithm and deep learning architecture significantly affects the accuracy of the models. The FT(2) embedding algorithm only performs well for CNN architecture, yielding relatively higher accuracy values across different n-gram versions of FastText model. The FT(3) embedding algorithm exhibits superior performance in most cases, except for the CNN model.. Notably, the FT(4) embedding algorithm consistently demonstrates the lowest accuracy scores across all models. In summary, the analysis of the provided table indicates that the selection of FastText n-gram embedding algorithm and deep learning architecture is crucial in achieving accurate results. The FT(2) embedding algorithm proves to be a reliable choice for only CNN model, while the FT(3) algorithm excels particularly in the BiLSTM model. Conversely, the FT(4) embedding algorithm generally performs less effectively.

**Table 7** Experiment Results of FastText Embedding Model including Various N-grams with Deep Learning Algorithms for Independent Dataset

| MODELS | ACC | SEN | SPE | MCC | AUC |
|---|---|---|---|---|---|
| FT(2)+CNN | 92.18 | 92.45 | 91.76 | 84.02 | 96.83 |
| FT(3)+CNN | 93.75 | 93.62 | 93.18 | 87.56 | 97.54 |
| FT(4)+CNN | 57.81 | 59.82 | 53.64 | 15.21 | 62.43 |
| FT(2)+LSTM | 87.50 | 88.24 | 86.36 | 73.86 | 91.57 |
| FT(3)+LSTM | 89.06 | 87.32 | 90.12 | 76.03 | 92.74 |
| FT(4)+LSTM | 62.33 | 56.34 | 59.28 | 16.07 | 58.89 |
| FT(2)+BiLSTM | **96.15** | 92.80 | 95.65 | 88.85 | 93.20 |
| FT(3)+BiLSTM | 93.85 | 92.87 | 94.63 | 87.41 | 96.85 |
| FT(4)+BiLSTM | 67.20 | 55.79 | 59.83 | 15.74 | 60.29 |

In Table 7, the accuracy results reveal important insights about the models that combine the FastText embedding algorithm with different deep learning architectures. The model utilizing the FT(2) embedding algorithm in conjunction with the CNN architecture demonstrates an accuracy value of 92.18%. The model incorporating the FT(3) embedding algorithm with the CNN architecture achieves a higher accuracy value of 93.75%. In contrast, the model employing the FT(4) embedding algorithm with the CNN architecture shows the lowest accuracy score of 57.81%. When considering the LSTM models, the combination of the FT(2) embedding algorithm with the LSTM architecture attains an accuracy value of 87.50%. Additionally, the model utilizing the FT(3) embedding algorithm with the LSTM architecture achieves a slightly higher accuracy value of 89.06%. However, the model incorporating the FT(4) embedding algorithm with the LSTM architecture demonstrates the lowest accuracy value of 62.33%. Turning attention to the BiLSTM models, the model utilizing the FT(3) embedding algorithm achieves an accuracy value of 93.85%. The combination of the FT(2) embedding algorithm with the BiLSTM architecture yields the highest accuracy value with 96.15%. Conversely, the model incorporating the FT(4) embedding algorithm with the BiLSTM architecture again demonstrates the lowest accuracy value of 67.20%.

As a result of Table 7, it becomes apparent that the selection of the FastText ngram embedding algorithm and deep learning architecture has a significant impact on the accuracy of the models. The FT(3) embedding algorithm consistently performs well, delivering higher accuracy values across CNN and LSTM models. Notably, the FT(2) algorithm performs the best classification success when combined with BiLSTM model, also exhibits competitive performance, particularly in the CNN and LSTM models. On the other hand, the FT(4) embedding algorithm consistently shows the lowest accuracy values across all models. As a result of Table 6 and Table 7, utilization of BiLSTM model notably boosts the classification performance when consolidated with convenient number of n-grams of FastText word embedding model.

**Table 8** Comparison with Literature Studies for ACPs250 Dataset

| Study | Model | Accuracy |
|---|---|---|
| [57] | Hajisharifi | 76.8 |
| [58] | IACP | 74.4 |
| [59] | ACPred-FL | 88.4 |
| [60] | CNN | 78.6 |
| [60] | DeepACP | 82.9 |
| [61] | ACPNet | 89.6 |
| **Proposed method** | FT(3)+BiLSTM | **92.50** |

In Table 8 and Table 9, the literature comparison is presented providing proposed models in the studies and accuracy scores. It is obviously observed that the proposed FT+BiLSTM framework exhibits remarkable experiment results with 92.50% of accuracy for ACPs250 dataset and 96.15% of accuracy for Independent dataset compared to the state-of-the-art studies.

The ROC (Receiver Operating Characteristic) curve serves as a metric for evaluating the performance of a classification model. The sharpness of the curve is associated with the model's classification capability and error rate. A sharper ROC curve typically represents a well-performing model. In this case, the model may exhibit high accuracy, low error rate, and proficient classification ability. This implies that the model's predictions provide a clearer distinction between the true classes. Conversely, a less sharp ROC curve may indicate lower classification performance for the model or a higher number of misclassifications in the confusion matrix. The sharpness of the ROC curve can depend on various factors, such as the features employed by the model, the training data, the complexity of the model, and the classification algorithm utilized. Optimizing or modifying these factors can influence the sharpness of the ROC curve. In conclusion, a sharper ROC curve signifies that the model demonstrates superior classification performance and provides a clearer differentiation between its predictions and the true classes.

**Table 9** Comparison with Literature Studies for Independent Dataset

| Study | Model | Accuracy |
|---|---|---|
| [62] | iACP-FSCM | 82.50 |
| [63] | mACPpred | 91.40 |
| [64] | ACPredStackL | 92.40 |
| [56] | iACP | 92.67 |
| [65] | Fm-Li | 93.61 |
| [66] | cACP-DeepGram | 94.02 |
| **Proposed method** | **FT(2)+BiLSTM** | **96.15** |

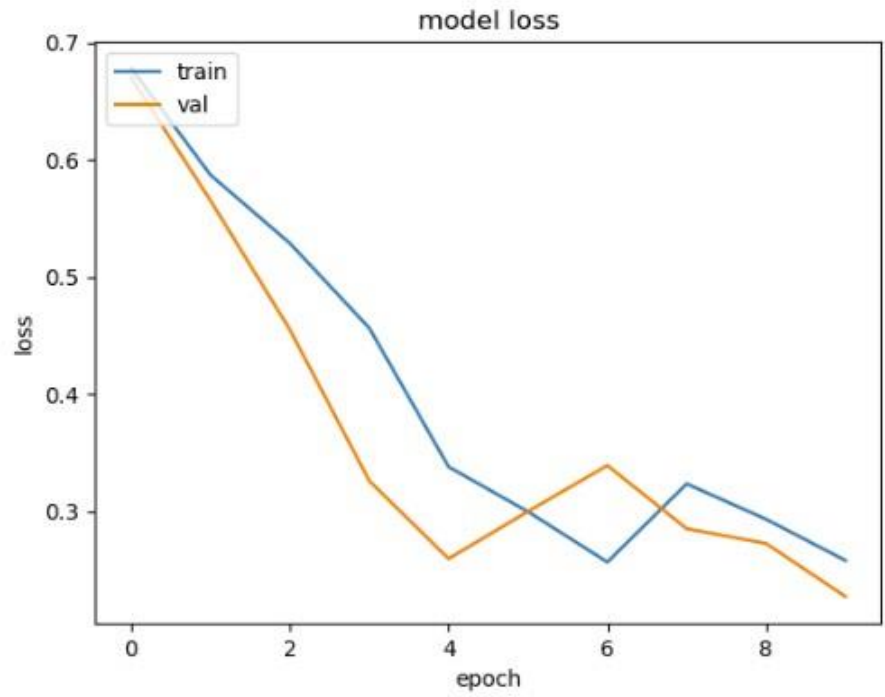

**Fig. 8** The Loss Graphic for Proposed FT(3)+BiLSTM Model of ACPs250 dataset

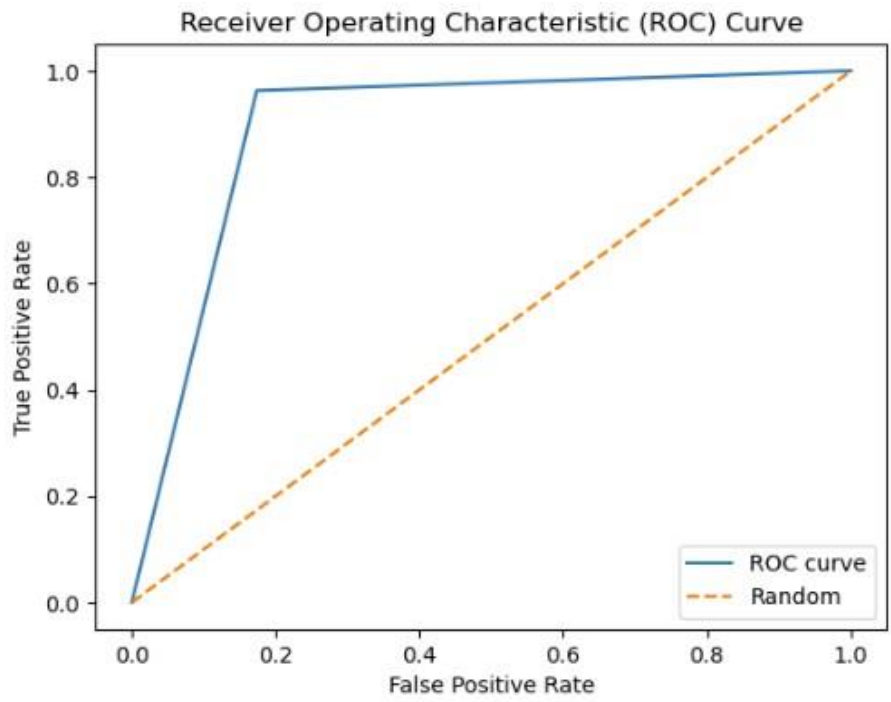

**Fig. 9** The ROC Curve Graphic for Proposed FT(3)+BiLSTM Model of ACPs250 dataset

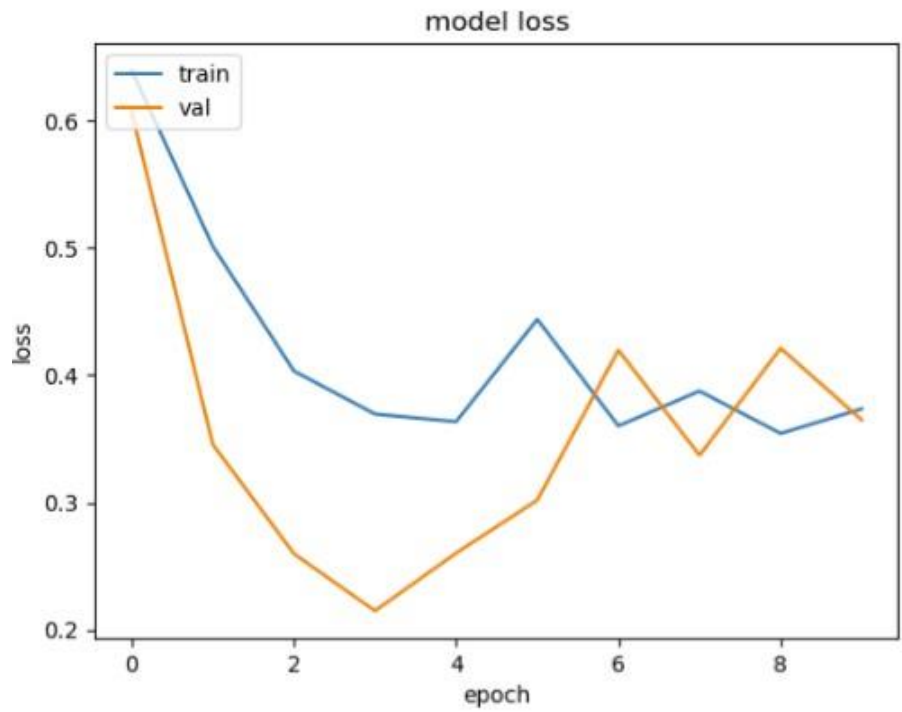

**Fig. 10** The Loss Graphic for Proposed FT(2)+BiLSTM Model of Independent dataset

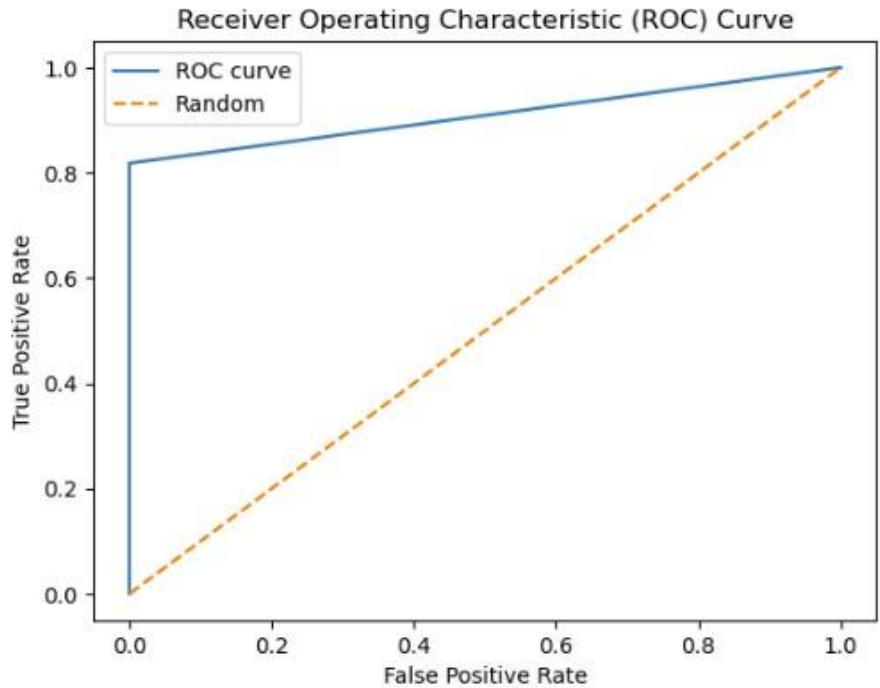

**Fig. 11** The ROC Curve Graphic for Proposed FT(2)+BiLSTM Model of Independent dataset

In Fig. 8 and Fig. 10, loss movements according to epoch size of training and validation sets are presented for ACPs250 and Independent datasets. In Fig. 8, earlystopping criteria, drop-out, and regularization techniques are applied to prevent the over-fitting problem. Thence, the training process is ended at epoch size 4. This means the best classification result is ensured with 92.50% of accuracy at epoch size 4 by handling over-fitting problem. It is very clear from the graph that the training and validation losses decrease in parallel with the number of trials until the 4th epoch size for ACPs250 dataset. In Fig. 10, the application of early-stopping criteria, drop-out, and regularization techniques is observed to mitigate the issue of over-fitting, as in Fig. 8. Consequently, the training process culminates at an epoch size of 3 guaranteeing the attainment of the most favorable classification outcome with an impressive accuracy rate of 96.15%. Notably, this achievement is attributed to the successful management of the over-fitting problem. The graphical representation vividly portrays a synchronized decline in both training and validation losses as the number of iterations progresses, effectively demonstrating the favorable convergence of these metrics up until the 3rd epoch size for the Independent dataset.

Instead of focusing solely on prediction rates, the researchers have employed Area under the Curve (AUC) as a metric, illustrated in Fig. 9 and Fig. 11. The ROC curve serves as the primary tool for assessing the model's stability and consistency. Without the ROC curve, there is no means to verify the model's consistent performance across multiple points. It is possible that the model excels in one parameter while performing poorly in others. In this study, the highest AUC value of 89.45% was achieved using 3-gram descriptors, as depicted in Fig. 9, while 2-gram descriptors exhibited an AUC value of 93.20% in Fig. 11.

# 6 Conclusions

This work introduces an efficient model for classifying anticancer peptides by combining word embedding techniques and deep learning models. The proposed framework utilizes Word2Vec and FastText models to extract peptide sequences, which are then fed into CNN, LSTM, and BiLSTM architectures. Extensive experiments are conducted on well-known datasets in this field, namely ACPs250 and Independent, to evaluate the performance of the proposed model. Throughout the study, skip-gram and CBOW versions of Word2Vec model and 2,3,and 4 gram versions of FastText word embedding algorithms are employed. Additionally, to enhance the reliability and performance of the models, a combination of techniques such as Dropout, pooling layer, and batch normalization are utilized. The experimental results demonstrate the effectiveness of the proposed approach, surpassing the success rate (89.6%) of stateof-the-art study [61] on the ACPs250 dataset and achieving a remarkable accuracy of 92.50%. Moreover, on the Independent dataset, the proposed model outperforms the leading study's [66] accuracy rate of 94.02% and achieves an impressive success rate of 96.15%.

The findings provide valuable insights into the strengths, limitations, and potential areas of improvement for deep learning-based approaches in ACP classification, informing future research in this field. The significance of this research lies in the development of efficient computational tools for accurate ACP classification, which can facilitate the discovery of novel anticancer therapeutics. The comparative analysis of deep learning models contributes to the existing knowledge in the field and shed light on their applicability and suitability for ACP classification. Ultimately, this research aims to enhance our understanding of ACPs and pave the way for the design of more effective and targeted anticancer treatments.

**Acknowledgments.** Not applicable.

**Author contributions.** Conceptualization, Zeynep Hilal Kilimci; methodology, Zeynep Hilal Kilimci; software, Onur Karakaya; validation, Onur Karakaya and Zeynep
Hilal Kilimci; formal analysis, Onur Karakaya and Zeynep Hilal Kilimci; investigation, Onur Karakaya and Zeynep Hilal Kilimci; data curation, Onur Karakaya; writing—original draft preparation, Onur Karakaya and Zeynep Hilal Kilimci; writing—review and editing, Zeynep Hilal Kilimci; visualization, Onur Karakaya; supervision, Zeynep Hilal Kilimci; project administration, Zeynep Hilal Kilimci; funding acquisition, Onur Karakaya.

**Funding.** This research received no external funding.

**Availability of data and materials.** We used publicly available datasets in our experiments. If necessary, we can share those datasets.